\documentclass[conference]{IEEEtran}
\IEEEoverridecommandlockouts
\usepackage{cite}
\usepackage{amsmath,amssymb,amsfonts}
\usepackage[utf8]{inputenc}
\usepackage{graphicx}
\usepackage{textcomp}
\usepackage{algorithm}
\usepackage{algpseudocode}
\usepackage{multirow}
\usepackage{makecell}
\usepackage{threeparttable}
\usepackage{graphicx} % for resizebox
\usepackage{tabularx}
\usepackage{booktabs}
\usepackage{enumitem}
\usepackage{float}
\setlist{nosep}
\UseRawInputEncoding
\usepackage{xcolor}
\def\BibTeX{{\rm B\kern-.05em{\sc i\kern-.025em b}\kern-.08em
    T\kern-.1667em\lower.7ex\hbox{E}\kern-.125emX}}

\usepackage{times}
\usepackage{amsmath,amssymb,amsfonts}
\usepackage{graphicx}
\usepackage{booktabs}
\usepackage{multirow}
\usepackage{array}
\usepackage{makecell}
\usepackage{url}
\usepackage{xcolor}
\usepackage{tikz}
\usepackage{pgfplots}
\pgfplotsset{compat=1.17}
\usepackage{caption}
\usepackage{subcaption}
\usepackage{dblfloatfix}
\usepackage{balance}
\usepackage{cite}
\usepackage{enumitem}
\usepackage{microtype}
\usepackage{rotating}
\usepackage{titlesec}
\renewcommand{\baselinestretch}{0.845}

\definecolor{darkblue}{RGB}{0,70,127}
\definecolor{accentblue}{RGB}{0,114,189}
\definecolor{promptgreen}{RGB}{0,128,96}
\definecolor{darkgray}{RGB}{80,80,80}

\newcommand{\bz}{\mathbf{z}}
\newcommand{\cE}{\mathcal{E}}
\newcommand{\cR}{\mathcal{R}}
\newcommand{\cH}{\mathcal{H}}
\newcommand{\Rbb}{\mathbb{R}}
\usepackage{xcolor}
\usepackage[normalem]{ulem}

\newif\ifshowcomments
\showcommentstrue
\ifshowcomments
    \newcommand{\fa}[1]{\textcolor{magenta}{[Fatemeh: #1]}}
       
    \newcommand{\ts}[1]{\textcolor{red}{[Tolunay: #1]}}
    \renewcommand{\mu}[1]{\textcolor{teal}{[Mohammad: #1]}}

    \newcommand{\fadel}[1]{\textcolor{blue}{\sout{#1}}}
    \newcommand{\tsdel}[1]{\textcolor{red}{\sout{#1}}}
    \newcommand{\mudel}[1]{\textcolor{teal}{\sout{#1}}}
\else
    \newcommand{\fa}[1]{}
    \newcommand{\ts}[1]{}
    \newcommand{\mu}[1]{}

    \newcommand{\fadel}[1]{}
    \newcommand{\tsdel}[1]{}
    \newcommand{\mudel}[1]{}
\fi

\begin{document}

\title{RFPrompt: Prompt-Based Expert Adaptation of the Large Wireless Model for Modulation Classification\thanks{This material is based upon work supported by the National Science Foundation under Grant Numbers CNS-2202972, CNS- 2318726, and CNS-2232048.}}

\author{\IEEEauthorblockN{ Md Raihan Uddin, Tolunay Seyfi, Fatemeh Afghah}
\IEEEauthorblockA{\textit{Department of Electrical and Computer Engineering, Clemson University, Clemson, SC, USA} \\
\{uddin2,tseyfi,fafghah\}@clemson.edu}
}

\maketitle

\begin{abstract}

Automatic modulation classification (AMC) in real-world deployments demands robustness to distribution shifts arising from hardware impairments, unseen propagation environments, and recording conditions never encountered during training. 
Although wireless foundation models offer a promising starting point for robust RF representation learning, an important open question is how to adapt them efficiently to out-of-distribution (OOD) downstream tasks without overwriting the structure learned during large-scale pre-training. In this paper, we investigate prompt-based adaptation as a general mechanism for OOD transfer in wireless foundation models. We propose RFPrompt, a parameter-efficient framework that introduces learnable deep prompt tokens while keeping the pretrained backbone frozen, enabling task-specific adaptation with minimal trainable parameters. We instantiate and evaluate this approach on the Large Wireless Model (LWM), a mixture-of-experts wireless foundation model, and study its behavior under both standard and OOD modulation-classification settings. Results show that prompt-based adaptation consistently improves robustness under distribution shift and limited supervision, particularly on real-world over-the-air IQ data, while preserving strong parameter efficiency. These findings suggest that prompt learning is a practical and effective strategy for adapting wireless foundation models to challenging downstream RF environments.
\end{abstract}


% ==========================================================
\section{Introduction}

The proliferation of heterogeneous wireless standards and increasingly contested spectrum has made automatic modulation classification (AMC) a critical function in cognitive radio, spectrum monitoring, and electronic intelligence systems. A deployed AMC system must classify incoming waveforms encoded as complex-valued in-phase/quadrature (IQ) samples across diverse signal-to-noise ratios (SNRs), hardware impairments, and propagation environments never seen during training. Deep learning has substantially advanced AMC performance. Convolutional networks and 
% \fa{define all acronyms the first time used} \yellow{\muadd{Long Short-Term Memory (LSTM)} }
Long Short-Term Memory (LSTM) applied to raw IQ or Short-Time Fourier Transform (STFT)-derived spectrograms achieves high accuracy on controlled benchmarks such as RadioML 2018~\cite{o2018over}. However, these models are trained from scratch on fixed datasets and generalize poorly to out-of-distribution (OOD) conditions, a fundamental barrier to real-world deployment where labeled radio frequency (RF) captures from every possible environment are unavailable. While techniques such as transfer learning, domain adaptation, and meta-learning have been explored to improve robustness to distribution shifts \cite{owfi2025adapt}.
% \fa{cite Ali Owfi's papers on AMC here}

% \fadel{Foundation models offer a compelling remedy. Pre-trained on large and diverse data, they encode general representations that transfer efficiently to downstream tasks with minimal task-specific supervision.} 
Recent wireless foundation models provide an alternative by learning transferable RF representations from large-scale, heterogeneous pre-training corpora, thereby reducing reliance on task-specific supervision and improving adaptation to downstream conditions that differ from the original training distribution.
% \fa{need a brief review of other WFM and citing them} 
IQFM~\cite{mashaal2026iqfm}, which pre-trains directly on raw multi-antenna IQ streams via contrastive self-supervised learning, and the Multimodal Wireless Foundation Model~\cite{aboulfotouh2025multimodal}, which jointly processes raw IQ and spectrogram representations through masked wireless modeling and demonstrates transfer across five heterogeneous physical-layer tasks. SpectrumFM~\cite{zhou2025spectrumfm}, which combines masked reconstruction and next-slot prediction for spectrum management tasks, including AMC and spectrum sensing, and RFGPT~\cite{zou2026rfgpt}, which injects RF spectrogram tokens into a decoder-only large language model (LLM) for natural-language RF reasoning. The Large Wireless Model (LWM)~\cite{alikhani2024large} 
% \fadel{extends this paradigm to the wireless domain: it}
pre-trains a spectrogram Vision Transformer (ViT)~\cite{dosovitskiy2020image} on 9.2 million IQ-derived spectrograms spanning LTE, WiFi, and 5G, augmented with a Mixture-of-Experts (MoE) backbone comprising a lightweight routing network and three protocol-specialized transformer encoders \cite{kim2026lwmspectro}. The core insight is that converting IQ signals to short-time Fourier transform (STFT) magnitude spectrogram frames pre-trains as a visual representation learning problem while grounding learned features in wireless propagation physics. LWM-Temporal~\cite{alikhani2026lwm} further extends the paradigm to spatiotemporal channel sequences via physics-informed sparse attention over the angle-delay-time domain.
% \fa{should add something here to say that the model is trained on synthetic DeepMIMO dataset and also only evaluated on synthetic dataset and as confirmed by our results, the performance of the model is very weak when tested on real-datasets, confirming the lack of OOD .} 
Although LWM is pretrained at scale, its development and evaluation are still centered on synthetic data generated via the DeepMIMO ray-tracing framework~\cite{alkhateeb2019deepmimo} or otherwise controlled data regimes, which do not fully capture the hardware variability and propagation complexity of real-world RF environments. Consequently, its robustness to true OOD downstream modulation classification remains unestablished. Our results confirm that the frozen LWM backbone provides insufficient out-of-distribution transfer performance on the real-world IQ dataset.

In this paper, we ask: \textit{how should the frozen LWM MoE backbone be adapted for downstream AMC, particularly under OOD conditions and limited labeled data?} 
We make the following contributions. 
% \fa{again, too tailored to LWM}
 \begin{itemize}[leftmargin=*]
%   \item \fadel{We propose \textbf{RFPrompt}, a deep prompt tuning framework tailored to the LWM MoE architecture, inserting expert-specific learnable prompt tokens at every transformer layer while keeping all backbone weights frozen.}
%   \item \fadel{We evaluate Frozen Expert, Partial Fine-Tuning (PFT), and RFPrompt on two IQ datasets of differing OOD severity under a data-scale sweep ($N \in \{100,\ldots,1600\}$) and a few-shot sweep ($N \in \{0, 2,\ldots,128\}$).}
  \item We formulate prompt-based adaptation as a parameter-efficient approach for transferring wireless foundation models to downstream AMC tasks under out-of-distribution conditions.
    \item We propose \textit{RFPrompt}, which adapts a frozen pretrained backbone through deep, learnable prompt tokens rather than backbone weight updates, allowing efficient specialization with minimal parameter overhead.
    \item Using LWM as a representative wireless foundation model, we show that this adaptation strategy is particularly effective for real-world OOD IQ classification and few-shot supervision, where pretrained features alone are insufficient and partial fine-tuning is less reliable.
    \item We provide empirical evidence that preserving the pretrained backbone while steering it through prompts offers a favorable trade-off between robustness, sample efficiency, and adaptation cost for wireless learning systems.
  \item RFPrompt closes over $79\%$ of the frozen-to-PFT gap using only $0.34\%$ of parameters and consistently outperforms conventional baselines trained from scratch across all shot budgets.
\end{itemize}

\begin{figure*}[t!]
  \centering
  \includegraphics[width=\linewidth]{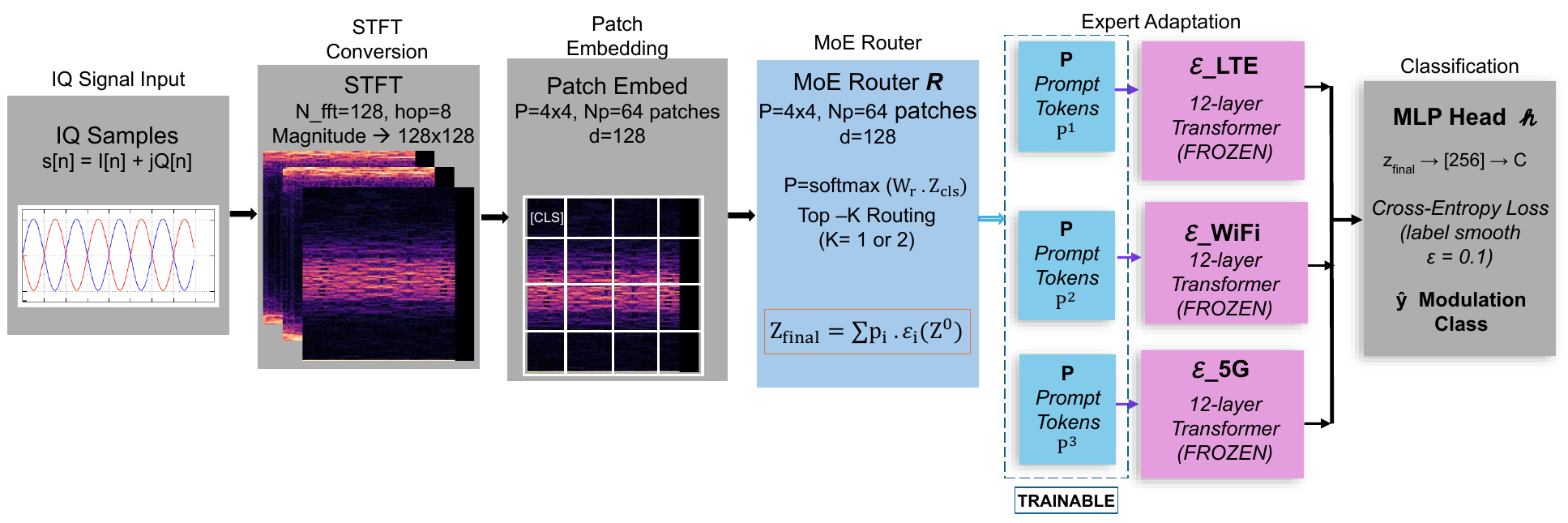}
  \caption{End-to-end overview of the proposed RFPrompt system. 
  % \fa{this figure could look better.}
  }
  \label{fig:system}
\end{figure*}

\section{Background and Related Work}

\subsection{Deep Learning for AMC.}
% \fa{same comment I had in the introduction, add a discussion on models handling OOD in traditional DL. cite more AMC papers from our group (Ali), ICC25 and ICNC } 
% O'Shea and Hoydis~\cite{o2017introduction} demonstrated that CNNs applied to IQ samples outperform handcrafted feature classifiers on RadioML benchmarks. Subsequent work explored residual networks~\cite{o2018over}, attention-augmented CNNs, and transformer-based spectrogram classifiers~\cite{shao2024iqformer}. A persistent limitation is narrow generalization: accuracy degrades sharply when SNR, hardware, or channel conditions are not represented in the training data.
Domain adaptation and adversarial robustness have been studied in tandem by Owfi et al.~\cite{owfi2025adapt}, proposes a two-phase framework combining meta-learning on adversarially perturbed source-domain samples with online domain adaptation to a shifted target environment. Authors also demonstrated~\cite{bamdad2025adaptive} that meta-learning-based adversarial training can generalize across a distribution of perturbation types, providing a strong non-foundation-model baseline for OOD-aware AMC. Hao et al.~\cite{zhang2020metaamc} address labeled-data scarcity through MAML with class-related mixup, generalizing to unseen modulation classes with as few as five labeled examples. RFPrompt instead inherits OOD invariance from a self-supervised wireless foundation model and acquires task-specific discriminability solely from prompt tokens, requiring neither adversarial augmentation nor episodic training.
\vspace{-4pt}
\subsection{Wireless Foundation Models.}
% \fa{same comment as Intro, mention more wireless foundation models} 
LWM~\cite{alikhani2024large} pre-trains a spectrogram transformer using masked spectrogram modeling and contrastive learning within a MoE framework, producing embeddings that exhibit strong SNR separation and modulation clustering in representation space. The MoE design, with three protocol-specialized experts and a learned router, enables protocol-aware feature extraction while sharing propagation modeling across experts \cite{kim2026lwmspectro}.  IQFM~\cite{mashaal2026iqfm} operates directly on raw multi-antenna IQ samples via contrastive self-supervised learning, achieving single-shot modulation classification with low-rank adaptation (LoRA). SpectrumFM~\cite{zhou2025spectrumfm} combines masked reconstruction and next-slot signal prediction for spectrum management tasks. RF-GPT~\cite{zou2026rfgpt} injects RF spectrogram tokens into a decoder-only LLM to enable natural-language reasoning over wireless signals. And, the Multimodal WFM~\cite{aboulfotouh2025multimodal} jointly pre-trains on raw IQ and spectrogram representations, demonstrating cross-task transfer across five physical-layer applications. Our work is the first to study prompt-based adaptation of LWM and to analyze the OOD transfer behavior of its MoE experts across varying supervision budgets.

\vspace{-4pt}
\subsection{Parameter-Efficient Fine-Tuning (PEFT) and Prompt Learning.}
Parameter-efficient fine-tuning methods adapt large pretrained models without updating the full backbone. Prompt tuning prepends soft trainable tokens to the input of a frozen language model and has since been extended to vision. Visual Prompt Tuning (VPT)~\cite{jia2022visual} inserts learnable tokens into a frozen ViT, with VPT-Deep, which places independent prompt tokens at every transformer layer, substantially outperforming shallow (input-only) prompting when the domain gap is large. Learning to Prompt (L2P)~\cite{wang2022learning} selects prompts per input from a shared pool via cosine similarity,
% \fadel{which remains shallow and cannot specialize across multiple parallel expert encoders}
but is still designed around a single shared backbone rather than multiple routed experts.
% \fa{define acronym} 
LoRA~\cite{hu2022lora} injects rank-decomposed weight updates into attention layers as a weight-space alternative, offering an alternative PEFT mechanism but one that alters backbone parameters rather than steering a fully frozen representation space.  
% \fadel{As summarized in Figure~\ref{fig:system}, RFPrompt is the only method that combines deep per-layer injection, expert-specific prompt assignment, and joint router training, making it uniquely suited to the LWM MoE structure.} 
% \fa{this is an irritating statement! "only" in what sense? across all domains? across all MoE models? across wireless? across AMC? We need to be specific about the novelty while not selling it as a simple applied model for a single model. See this and revise as needed.} 
Existing prompt-learning methods were developed primarily for single-backbone language or vision transformers, whereas LWM uses a routed mixture-of-experts architecture with protocol-specialized encoders. This creates a different adaptation problem: the method must preserve expert specialization, support layer-wise modulation of frozen representations, and remain compatible with routing. As summarized in Figure~\ref{fig:system}, RFPrompt addresses this setting by jointly optimizing expert-specific deep prompts with the router and classifier.

% \begin{table}[t!]
%   \centering
%   \caption{Comparison of prompt-based adaptation methods.
%     ``Expert-specific'' = independent prompts per encoder.
%     ``MoE-aware'' = joint training with the routing network.}
%   \label{tab:prompt_compare}
%   \setlength{\tabcolsep}{3pt}
%   \renewcommand{\arraystretch}{1.05}
%   \begin{tabular}{lcccc}
%     \toprule
%     \textbf{Method} & \textbf{Depth}
%       & \makecell{\textbf{Expert}\\\textbf{specific}}
%       & \makecell{\textbf{MoE}\\\textbf{aware}}
%       & \textbf{Domain} \\
%     \midrule
%     NLP PT~\cite{lester2021power}
%       & Shallow & \texttimes & \texttimes & NLP \\
%     VPT-Deep~\cite{jia2022visual}
%       & Deep    & \texttimes & \texttimes & Vision \\
%     L2P~\cite{wang2022learning}
%       & Shallow & \texttimes & \texttimes & Vision \\
%     \textbf{RFPrompt}
%       & \textbf{Deep}
%       & \textbf{\checkmark}
%       & \textbf{\checkmark}
%       & \textbf{Wireless} \\
%     \bottomrule
%   \end{tabular}
% \end{table}

\section{System Model}

Fig.~\ref{fig:system} illustrates the end-to-end pipeline. A received RF signal is represented as $N_s{=}1024$ complex IQ samples $s[n] = I[n] + jQ[n]$, $n = 0,\ldots,N_s{-}1$, zero-padded or truncated as needed, where each observation carries a modulation label $y \in \{1,\ldots,C\}$. 
% \fa{have not talked bout the datasets yet, so this does not make sense for reviewers, add a bullet item or statement to intro describing the real datasets} , a technology tag $\tau \in \{\text{LTE},\text{WiFi},\text{5G}\}$.
To maintain strict compatibility with pre-training statistics, the IQ signal is converted to a magnitude spectrogram via the STFT,
\begin{equation}
  \mathbf{X}[k,m] = \left|\!
    \sum_{n=0}^{N_{\mathrm{fft}}-1}
    s[n]\cdot w[n{-}mH]\cdot e^{-j2\pi kn/N_{\mathrm{fft}}}
  \!\right|,
  \label{eq:stft}
\end{equation}
with Hann window $w[\cdot]$, $N_{\mathrm{fft}}{=}128$, and hop length $H{=}8$ samples between adjacent STFT frames. We use no implicit zero-padding before framing, and each frame starts at the true sample index $mH$. The magnitude spectrogram is then zero-padded or cropped to $128{\times}128$. No normalization beyond defaults is applied, preserving the statistics expected by the pre-trained patch projection layers.

The backbone processes $\mathbf{X}$ as a sequence of patch tokens. The $128{\times}128$ spectrogram is partitioned into $N_p{=}64$ non-overlapping $4{\times}4$ patches, each linearly projected to $d{=}128$ dimensions. A learnable \texttt{[CLS]} token is prepended, and learnable absolute positional embeddings are added, giving $\mathbf{Z}^0\in\Rbb^{(N_p+1)\times d}$. Each of the $L{=}12$ transformer layers then applies a pre-norm residual block.
% \begin{align}
%   \mathbf{A}^l &=
%     \text{MHSA}\!\bigl(\text{LN}(\mathbf{Z}^{l-1})\bigr)
%     + \mathbf{Z}^{l-1}, \\
%   \mathbf{Z}^l &=
%     \text{FFN}\!\bigl(\text{LN}(\mathbf{A}^l)\bigr) + \mathbf{A}^l,
% \end{align}
MHSA denotes multi-head self-attention, FFN denotes a position-wise two-layer feed-forward network, and GELU is the Gaussian Error Linear Unit activation. We use $H_{\text{attn}}{=}4$ attention heads and FFN hidden dimension $4d$, yielding ${\approx}1.6\text{M}$ parameters per expert.

The three independent copies of this encoder are protocol-specialized experts
$\cE=\{\cE_{\text{LTE}},\cE_{\text{WiFi}},\cE_{\text{5G}}\}$.
A lightweight router maps the shared \texttt{[CLS]} embedding to a soft distribution over experts,
% \begin{equation}
%   \mathbf{p} = \softmax(\mathbf{W}_r\bz_{\texttt{[CLS]}})
%     \in \Rbb^3,
%   \label{eq:router}
% \end{equation}
and top-$k$ routing ($k{=}1$ or $2$) produces the final representation as their weighted combination.
% \begin{equation}
%   \bz_{\text{final}} =
%     \sum_{i\in\mathcal{S}_k} p_i\cdot\cE_i(\mathbf{Z}^0)
%     \in \Rbb^d.
%   \label{eq:moe}
% \end{equation}
The aggregated embedding is classified by a two-layer MLP head,
\begin{equation}
  \hat{y} = \mathbf{W}_2\,
    \text{GELU}\!\bigl(\text{LN}(
    \mathbf{W}_1\bz_{\text{final}}{+}\mathbf{b}_1)\bigr)
    + \mathbf{b}_2,
\end{equation}
with hidden dimension $256$ and $C$ output classes, trained with cross-entropy loss and label smoothing $\epsilon{=}0.1$.

\section{Proposed RFPrompt}
\label{sec:rfprompt}

% \begin{figure*}[t!]
%   \centering
%   \includegraphics[width=\linewidth]{fig2_rfp.pdf}
%   \caption{Comparison of prompt-based adaptation paradigms.
%     (a)~NLP prompt tuning \cite{lester2021power}: soft prompts at input only.
%     (b)~VPT-Deep \cite{jia2022visual}: independent prompt tokens at every ViT layer.
%     (c)~L2P \cite{wang2022learning}: input-conditioned selection from a shared pool.
%     (d)~\textbf{RFPrompt}: per-expert, per-layer deep prompts with
%     jointly trained router.}
%   \label{fig:prompt_compare}
% \end{figure*}

\vspace{-4pt}
\subsection{Motivation}
% \fadel{Three properties of LWM make existing PEFT methods inadequate for this setting: (i)~The three experts occupy distinct representation subspaces tuned to ; a single shared prompt cannot align all three;} 
Three properties of LWM motivate a more tailored PEFT strategy for this setting. First, its three experts are protocol-specialized (e.g., LTE guard intervals, WiFi OFDM pilots, and 5G NR slot structures) and therefore likely operate in different representation subspaces, as a result, a single shared prompt may be insufficient to capture the distinct adaptation needs of LTE, WiFi, and 5G experts. 
% \fadel{(ii)~Modulation-discriminative features, including constellation density and symbol-rate harmonics, emerge in intermediate and late transformer layers, so input-only prompts attenuate before reaching them; and (iii)~All backbone weights must remain frozen to preserve the protocol specialization that makes MoE routing meaningful \cite{mo2026pmoe, le2026revisit}} \fa{why would we say that? this is not for general cases and even here our results prove it otherwise as our PFT improves the result} . 
% \textcolor{red}{RFPrompt satisfies all three requirements simultaneously.} 
% \fa{always avoid using such sentences}
Second, modulation-relevant cues such as constellation-density patterns and symbol-rate harmonics are not necessarily isolated to the input layer, suggesting that layer-wise prompting may be better suited than input-only prompting for downstream AMC adaptation. Third, because LWM relies on routed expert specialization, we focus on a lightweight adaptation regime that preserves the pretrained expert backbone and avoids full backbone re-optimization \cite{mo2026pmoe, le2026revisit}. These considerations motivate RFPrompt, which uses expert-specific deep prompts to adapt the model while retaining the original MoE structure.

% \subsection{Per-Expert Deep Prompt Tokens}

% For each expert $\cE_i$, $i\in\{\text{LTE},\text{WiFi},\text{5G}\}$, and each layer $l\in\{1,\ldots,L\}$, we define $M$ independent learnable prompt tokens:
% \begin{equation}
%   \mathbf{P}^l_i \in \Rbb^{M\times d},
%   \quad
%   \mathbf{P}^l_i \sim \mathcal{N}(0,\,0.02^2\,\mathbf{I}),
%   \label{eq:rfprompt_init}
% \end{equation} \fa{why this specific standard deviation, I suggest using a parameter here and mentioning the exact number in evaluation results}
% initialized with a small variance to avoid disturbing frozen attention statistics at the start of training.

% \subsection{Layer-wise Prompt Injection}

% At each layer $l$ of expert $\cE_i$, prompt tokens are prepended
% before MHSA:
% \begin{equation}
%   \tilde{\mathbf{Z}}^l_i =
%     \bigl[\mathbf{P}^l_i\;;\;\mathbf{Z}^{l-1}_i\bigr]
%     \in\Rbb^{(M+N_p+1)\times d}.
%   \label{eq:rfprompt_concat}
% \end{equation} \fa{make sure all parameters are defined}
% The frozen layer processes the augmented sequence with standard pre-norm residual computation, and the $M$ prompt output positions are discarded before the next layer:
% \begin{equation}
%   \mathbf{Z}^l_i =
%     \tilde{\mathbf{Z}}^{l\prime}_i[M{:},]
%     \in\Rbb^{(N_p+1)\times d},
%   \label{eq:rfprompt_discard}
% \end{equation}
% keeping the downstream sequence length and all backbone weight statistics unchanged.
\vspace{-4pt}
\subsection{Per-Expert Deep Prompt Tokens}
For each expert $\cE_i$, $i\in\{\text{LTE},\text{WiFi},\text{5G}\}$,
and each transformer layer $l\in\{1,\ldots,L\}$, we introduce $M$
independent learnable prompt token vectors:
\begin{equation}
  \mathbf{P}^l_i \in \Rbb^{M\times d},
  \quad
  \mathbf{P}^l_i \sim \mathcal{N}(0,\,\sigma^2\mathbf{I}),
  \label{eq:rfprompt_init}
\end{equation}
where $M$ is the prompt length, $d$ is the token embedding dimension, and $\sigma$ is a small initialization scale chosen to keep the initial prompt tokens close to zero so that the frozen attention statistics are not disturbed at the start of training. The specific values of $M$ and $\sigma$ are determined.

\vspace{-4pt}
\subsection{Layer-wise Prompt Injection}
At each layer $l$ of expert $\cE_i$, the $M$ prompt tokens are prepended to the current token sequence before multi-head
self-attention:
\begin{equation}
  \tilde{\mathbf{Z}}^l_i =
    \bigl[\mathbf{P}^l_i\;;\;\mathbf{Z}^{l-1}_i\bigr]
    \in\Rbb^{(M+N_p+1)\times d},
  \label{eq:rfprompt_concat}
\end{equation}
where $\mathbf{Z}^{l-1}_i \in \Rbb^{(N_p+1)\times d}$ is the token sequence passed from layer $l{-}1$ (comprising the \texttt{[CLS]} token and $N_p$ patch tokens), and $\tilde{\mathbf{Z}}^l_i$ is the augmented sequence of length $M+N_p+1$. The frozen transformer layer processes $\tilde{\mathbf{Z}}^l_i$ through its standard pre-norm residual blocks (MHSA and FFN), producing an output $\tilde{\mathbf{Z}}^{l\prime}_i \in \Rbb^{(M+N_p+1)\times d}$. The first $M$ positions, corresponding to prompt token outputs, are then discarded before passing to the next layer:
\begin{equation}
  \mathbf{Z}^l_i =
    \tilde{\mathbf{Z}}^{l\prime}_i[M{:},\,:]
    \in\Rbb^{(N_p+1)\times d},
  \label{eq:rfprompt_discard}
\end{equation}
where the slice $[M{:},\,:]$ selects rows $M$ through $M+N_p$,
recovering the original sequence length $(N_p+1)$. This ensures that the token dimensionality seen by all downstream layers matches the expectations of the frozen backbone and that backbone weight statistics remain unaffected.

\vspace{-4pt}
\subsection{Attention Steering Mechanism}
Standard MHSA computes attention over all $M{+}N_p{+}1$ tokens including the prompt tokens:
\begin{equation}
  \mathrm{Attn}(\mathbf{Q},\mathbf{K},\mathbf{V})
    = \mathrm{softmax}\!\!\left(
      \frac{\mathbf{Q}\mathbf{K}^\top}{\sqrt{d_k}}
    \right)\!\mathbf{V}.
  \label{eq:attn}
\end{equation}
The $M$ prompt tokens contribute additional rows to $\mathbf{K}$ and
$\mathbf{V}$.
For patch token $j$, its attention score toward prompt token $p$ is:
\begin{equation}
  a_{j,p} =
    \frac{\mathbf{q}_j^\top\mathbf{k}_p}{\sqrt{d_k}},
  \quad
  \mathbf{k}_p = \mathbf{W}_K\mathbf{P}^l_i[p,:].
  \label{eq:attn_score}
\end{equation}
Through gradient descent, $\mathbf{P}^l_i$ adjusts so that the resulting attention distribution amplifies focus on modulation-discriminative time-frequency bins, such as symbol-rate harmonics and constellation density patterns. This per-layer steering is not achievable by shallow methods: a prompt injected only at layer 0 is mixed and attenuated by the first FFN, leaving layers 2 through 12 uninformed of the adaptation direction.

\subsection{Joint Router Training and Optimization}
After all $L$ layers, the \texttt{[CLS]} embedding of each expert is combined via the router as in~(\ref{eq:moe}). The router $\theta_\cR$ is trained jointly with the prompt tokens. Because prompts are expert-specific, each expert's output is steered toward a distinct modulation-relevant subspace, providing the router with richer inputs than frozen-expert embeddings alone.

All backbone parameters are frozen; the full trainable set is:
\begin{equation}
  \Phi = \{\mathbf{P}^l_i\}\cup\theta_\cR\cup\theta_\cH.
\end{equation}
Training minimizes cross-entropy with label smoothing $\epsilon{=}0.1$:
\begin{equation}
  \mathcal{L}(\Phi) =
    -\sum_{c=1}^{C}\tilde{y}_c\log\hat{y}_c,
  \quad
  \tilde{y}_c =
    (1{-}\epsilon)\mathbf{1}[c{=}y] + \tfrac{\epsilon}{C}.
\end{equation}

\subsection{Parameter Efficiency}

% With $M{=}16$, $L{=}12$, three experts, $d{=}128$:
% \begin{equation}
%   |\mathbf{P}|=16\times12\times3\times128=73{,}728.
% \end{equation}
Including router (${\approx}16\text{K}$) and head (${\approx}165\text{K}$), RFPrompt trains ${\approx}255\text{K}$ parameters, i.e.\ ${\approx}0.34\%$ of the total $4.8\text{M}$ LWM MoE parameters (Table~\ref{tab:params}).

\begin{table}[!t]
  \centering
  \vspace{5pt}
  \caption{Trainable parameter count per adaptation regime.}
  \label{tab:params}
  \vspace{-4pt}
  \setlength{\tabcolsep}{4pt}
  \begin{tabular}{lccc}
    \toprule
    \textbf{Regime} & \textbf{Backbone} & \textbf{Prompts}
      & \textbf{Total (\%)} \\
    \midrule
    Frozen Expert    & 0                       & --       & ${\approx}0.37\%$ \\
    PFT (last 2 blks)& ${\approx}800\text{K}$  & --       & ${\approx}17\%$   \\
    \textbf{RFPrompt}& 0                       & 73{,}728 & ${\approx}0.34\%$ \\
    \bottomrule
  \end{tabular}
  \vspace{-4pt}
\end{table}

\section{Experimental Setup}

\subsection{Datasets}
We evaluate on two IQ datasets differing in distributional proximity to LWM pre-training. The \textit{IEEE Dataport IQ} dataset contains baseband IQ samples from LTE, 5G NR, and WiFi transmitters captured via SDR hardware under controlled conditions; we use five rate-and-technology classes and convert raw IQ to magnitude spectrograms~\cite{girmay2023ieeeiq}. This dataset represents a low distribution shift relative to LWM pre-training and serves as the in-distribution benchmark. The \textit{Real-World IQ} dataset consists of over-the-air captures in realistic environments containing multipath fading, co-channel interference, and hardware imperfections; we use seven modulation and traffic classes~\cite{belousov2026realworldiq}. Samples contain only modulation class labels, with no technology metadata, so the router must infer protocol assignments without supervision, making this the primary OOD benchmark.
% \fa{how about RML18?}

\subsection{Preprocessing}
We replicate the exact STFT configuration of the official LWM codebase~\cite{alikhani2024large}: 1024-sample IQ windows, $N_{\mathrm{fft}}{=}128$, $H{=}8$, Hann window, magnitude as \texttt{float32}, resized to $128{\times}128$. No normalization beyond LWM defaults is applied.

\subsection{Data splits and supervision axes}
A stratified 70/15/15 train/validation/test split is applied; validation and test sets remain uncapped throughout. 
% \fadel{\textit{Stage~A} caps the training set at $N \in \{100, 200, 400, 800, 1600\}$ samples per class. \textit{Stage~B} caps it at $N \in \{0, 2, 4, 8, 16, 32, 64, 128\}$ shots per class, where $N{=}0$ is zero-shot inference.} \fa{using the same notation N for both stages is confusing, let's call N in stage B, as K shot. I suggest changing the order of 'data splits and supervision section and adaptation regime section and add something like following to explain these scenarios} 
We distinguish between two supervision regimes. In Stage~A, $N$ denotes a per-class training cap used to study supervised target-domain data scaling. This setting evaluates how adaptation methods behave as more labeled downstream data become available. In Stage~B, $K$ denotes the few-shot support size, i.e., the number of labeled examples per class available for target-task adaptation. This corresponds to the standard $C$-way $K$-shot terminology used in few-shot learning. Unlike meta-learning methods such as Model-Agnostic Meta-Learning (MAML), First-Order MAML (FOMAML), and Reptile, which learn an adaptation rule from a distribution of episodic tasks, RFPrompt performs parameter-efficient adaptation of a pretrained wireless foundation model to a single downstream RF task by learning prompts, the router, and the classifier parameters while keeping the backbone frozen. Thus, Stage~B evaluates few-shot target-domain adaptation rather than episodic meta-learning.

\subsection{Adaptation regimes}
The router $\cR$ and head $\cH$ are trainable in all regimes. \textit{Frozen Expert (FE)}: all expert backbone weights $\theta_\cE$ frozen; only $\cR$ and $\cH$ are optimized. \textit{Partial Fine-Tuning (PFT)}: the final two transformer blocks (layers~10--11) of each expert are unfrozen alongside $\cR$ and $\cH$; earlier layers are kept frozen to reduce forgetting. \textit{RFPrompt (Ours)}: expert-specific deep prompt tokens (Section~\ref{sec:rfprompt}); all backbone weights frozen. We further compare against ResNet18~\cite{he2016deep}, EfficientNet-B0~\cite{tan2019efficientnet}, MobileNetV3~\cite{howard2019searching}, a spectrogram CNN~\cite{wu2021radio}, MAML~\cite{finn2017model}, FOMAML, Reptile~\cite{nichol2018reptile}, and Transfer\_RML16 (pre-trained on RadioML 2016.10a), all trained from scratch. 
% \ts{We also include SpectrumFM (pre-trained on RadioML 2018) as a wireless foundation-model baseline and adapt it to each downstream dataset using partial fine-tuning with a task-specific classification head.} 
We also include SpectrumFM (pre-trained on RadioML 2018) as a wireless foundation model baseline and adapt it to each downstream dataset using partial fine-tuning with a task-specific classification head.
% \fa{define MAML, and FOMAML acronyms}.

\subsection{Training configuration}
All runs use AdamW~\cite{loshchilov2017decoupled} ($\lambda{=}0.01$) with differential learning rates: $\eta_b{=}10^{-5}$ for PFT backbone layers and $\eta_h{=}10^{-3}$ for router, head, and prompt parameters. A cosine annealing schedule with 5-epoch linear warm-up is applied; the router undergoes a dedicated 2-epoch warm-up before expert adaptation begins. Maximum training is 100 epochs with early stopping, and a batch size of 32 is used on 12 GB GPUs.

\section{Results and Discussion}
\label{sec:results}

\subsection{Data-Scale Sweep (Stage~A)}

Table~\ref{tab:stageA} reports test accuracy across all training caps
and baselines for both datasets.

\begin{table}[t!]
  \centering
  \vspace{5pt}
  \caption{Stage A (Data-Scale Sweep): Test Accuracy vs. Training Cap $N$. Tr: Transfer, FML: FOMAML, SFM: SpectrumFM.}
  \label{tab:stageA}
  \vspace{-4pt}
  \setlength{\tabcolsep}{4.2pt}
  \renewcommand{\arraystretch}{1.08}
  \scriptsize
  \begin{tabular}{@{}lccccccc@{}}
    \toprule
    \multicolumn{8}{c}{\textbf{IEEE Dataport IQ}} \\
    \midrule
    $N$ & \textbf{Frz} & \textbf{PFT} & \textbf{RFP} & \textbf{Tr} & \textbf{MML} & \textbf{FML} & \textbf{SFM} \\
    \midrule
    100  & .38 & .59 & \textbf{.70} & .62 & .66 & .69 & .39 \\
    200  & .38 & .67 & .71 & .74 & \textbf{.75} & .74 & .55 \\
    400  & .39 & .68 & .74 & .80 & \textbf{.80} & .79 & .66 \\
    800  & .44 & .58 & .70 & .82 & \textbf{.82} & .81 & .80 \\
    1600 & .49 & .72 & .72 & .84 & .83 & .83 & \textbf{.86} \\
    \midrule
    \multicolumn{8}{c}{\textbf{Real-World IQ}} \\
    \midrule
    $N$ & \textbf{Frz} & \textbf{PFT} & \textbf{RFP} & \textbf{Tr} & \textbf{MML} & \textbf{FML} & \textbf{SFM} \\
    \midrule
    100  & .68 & \textbf{.70} & .69 & .15 & .31 & .25 & .42 \\
    200  & .72 & .71 & \textbf{.79} & .15 & .36 & .36 & .57 \\
    400  & .66 & .76 & \textbf{.82} & .16 & .39 & .42 & .63 \\
    800  & .69 & .75 & \textbf{.80} & .16 & .53 & .56 & .65 \\
    1600 & .69 & .77 & \textbf{.80} & .16 & .64 & .64 & .73 \\
    \bottomrule
  \end{tabular}
  \vspace{-4pt}
\end{table}

\textbf{IEEE Dataport (in-distribution).} RFPrompt achieves the highest or joint-highest accuracy at every training cap up to $N{=}800$, reaching $0.74$ at $N{=}400$ while updating only $0.34\%$ of parameters, versus PFT at $17\%$. The Frozen Expert baseline grows slowly from $0.38$ to $0.49$, confirming that protocol-tuned pre-trained features still require task-specific guidance to reach competitive classification accuracy. 
Meta-learning methods (MAML, FOMAML) become increasingly competitive as $N$ grows, surpassing RFPrompt from $N{=}200$ onward by leveraging a diverse task distribution during meta-training. Transfer\_RML16 follows a similar trend, benefiting from pre-trained features on RadioML~2016.10a that are moderately aligned with the IEEE Dataport distribution. SpectrumFM also improves with additional target-domain supervision, rising from $0.39$ at $N{=}100$ to $0.86$ at $N{=}1600$, achieving the best accuracy among the compared methods.

\begin{table*}[!t]
  \centering
  \vspace{5pt}
  \caption{Stage~B (Few-Shot Sweep): classification accuracy (\%) vs.\ shot budget across all methods. CNN baselines and meta-learning methods are trained from scratch at each shot level.}
  \label{tab:stageB}
  \vspace{-4pt}
  \setlength{\tabcolsep}{5.0pt}
  \renewcommand{\arraystretch}{1.05}
  \scriptsize
  \begin{tabular}{@{}l | cccccccc | cccccccc@{}}
    \toprule
    & \multicolumn{8}{c|}{\textbf{IEEE Dataport IQ}} & \multicolumn{8}{c}{\textbf{RW-IQ (OOD)}} \\
    \midrule
    \textbf{Method}
    & \textbf{0} & \textbf{2} & \textbf{4} & \textbf{8}
    & \textbf{16} & \textbf{32} & \textbf{64} & \textbf{128}
    & \textbf{0} & \textbf{2} & \textbf{4} & \textbf{8}
    & \textbf{16} & \textbf{32} & \textbf{64} & \textbf{128} \\
    \midrule
    \textbf{RFPrompt (Ours)}
    & 20.0 & \textbf{37.4} & \textbf{47.0} & \textbf{50.2}
    & \textbf{61.0} & 58.8 & \textbf{65.2} & \textbf{73.0}
    & 14.3 & \textbf{52.6} & \textbf{60.6} & \textbf{69.4}
    & \textbf{69.8} & \textbf{75.9} & \textbf{80.4} & \textbf{83.3} \\
    \midrule
    ResNet18
    & 19.8 & 24.4 & 29.2 & 33.6
    & 22.8 & 59.0 & 59.0 & 72.8
    & 6.3 & 13.7 & 18.3 & 28.9
    & 40.6 & 40.9 & 53.0 & 60.7 \\
    EffNet-B0
    & \textbf{20.2} & 28.2 & 28.6 & 26.2
    & 40.0 & \textbf{59.6} & 61.0 & 64.8
    & 10.7 & 12.9 & 20.3 & 21.4
    & 14.9 & 25.7 & 42.4 & 39.9 \\
    MobileNetV3
    & 17.4 & 19.8 & 27.0 & 27.0
    & 19.8 & 24.2 & 24.8 & 24.0
    & 14.3 & 14.6 & 14.6 & 20.7
    & 20.1 & 14.3 & 21.3 & 15.6 \\
    CNN
    & 20.0 & 20.0 & 20.0 & 34.4
    & 50.4 & 50.8 & 58.4 & 69.2
    & 14.3 & 15.6 & 14.4 & 14.3
    & 21.3 & 22.9 & 13.1 & 44.0 \\
    \midrule
    Transfer
    & 20.7 & 32.6 & 31.5 & 34.3
    & 41.3 & 48.1 & 60.8 & 71.5
    & 14.5 & 15.5 & 15.5 & 15.5
    & 15.5 & 15.5 & 15.4 & 35.0 \\
    MAML
    & 20.0 & 33.1 & 25.2 & 38.7
    & 41.0 & 52.9 & 63.8 & 72.4
    & 14.5 & 16.4 & 15.6 & 16.3
    & 15.5 & 20.4 & 29.4 & 30.5 \\
    FOMAML
    & 2.9 & 27.7 & 29.6 & 31.9
    & 44.4 & 56.2 & 62.5 & 71.6
    & 14.5 & 14.6 & 15.7 & 16.0
    & 15.4 & 21.0 & 20.2 & 32.9 \\
    Reptile
    & 14.3 & 17.8 & 19.1 & 20.3
    & 22.9 & 25.1 & 26.0 & 25.3
    & \textbf{15.4} & 16.9 & 18.3 & 19.4
    & 20.8 & 20.5 & 22.7 & 20.8 \\
    SpectrumFM
    & 19.7 & 23.8 & 27.9 & 24.5
    & 29.1 & 29.3 & 44.6 & 47.0
    & 9.1 & 26.2 & 24.5 & 23.7
    & 24.4 & 31.1 & 38.3 & 48.1 \\
    \bottomrule
  \end{tabular}
  \vspace{-4pt}
\end{table*}

\textbf{Real-World IQ (OOD).}
% The OOD dataset reveals substantially sharper differentiation. At $N{=}100$, all LWM-based regimes cluster within $\{0.68, 0.70, 0.69\}$, confirming that fewer than 100 labeled samples per class are insufficient to specify an effective adaptation direction regardless of method. Meta-learning baselines collapse on this dataset, with Transfer and MAML achieving at most $0.64$ accuracy even at $N{=}1600$, indicating that their meta-training distributions do not transfer to the hardware and channel conditions of the Real-World IQ captures. RFPrompt surpasses all baselines from $N{=}200$ onward, peaking at $0.82$ at $N{=}400$, a gain of $+0.16$ over Frozen Expert and $+0.06$ over PFT. The plateau from $N{=}400$ to $N{=}1600$ ($0.82\to0.80$) suggests prompt capacity saturates at the current $M{=}16$ setting; increasing $M$ or applying prompting to additional layers may extend the performance ceiling.

The OOD dataset reveals substantially sharper differentiation. At $N{=}100$, all LWM-based regimes cluster within $\{0.68, 0.70, 0.69\}$, confirming that fewer than 100 labeled samples per class are insufficient to specify an effective adaptation direction regardless of method. Meta-learning baselines collapse on this dataset, with Transfer and MAML achieving at most $0.64$ accuracy even at $N{=}1600$, indicating that their meta-training distributions do not transfer to the hardware and channel conditions of the Real-World IQ captures. SpectrumFM improves with additional target-domain supervision, increasing from $0.42$ at $N{=}100$ to $0.73$ at $N{=}1600$, but remains below RFPrompt across the reported training caps. RFPrompt surpasses all baselines from $N{=}200$ onward, peaking at $0.82$ at $N{=}400$, a gain of $+0.16$ over Frozen Expert and $+0.06$ over PFT. The plateau from $N{=}400$ to $N{=}1600$ ($0.82\to0.80$) suggests prompt capacity saturates at the current $M{=}16$ setting; increasing $M$ or applying prompting to additional layers may extend the performance ceiling.

\textbf{Attention steering explains the OOD advantage.}
The frozen LWM experts encode time-frequency features tailored to LTE, WiFi, and 5G waveform structures. For modulation classification, discriminative information is concentrated in a small subset of those features, primarily constellation density patterns and symbol-rate harmonics. Via~(\ref{eq:attn_score}), the learned prompt tokens reshape the layer-wise attention weight matrix to amplify focus on these bins and suppress channel artifacts such as Doppler smearing and multipath spreading that are prevalent in OOD captures. Because this steering operates on frozen representations without altering backbone weights, RFPrompt converges more reliably than PFT under limited OOD supervision, and more robustly than meta-learning methods whose task distribution does not encompass real-world hardware impairments.

\vspace{-4pt}
\subsection{Few-Shot Sweep and Baseline Comparison (Stage~B)}
Table~\ref{tab:stageB} reports accuracy across shot levels $K \in \{0, 2, \ldots, 128\}$ for all methods. The $K{=}0$ row reflects zero-shot performance inference with no downstream labels, serving as a lower bound on pre-trained feature transferability.

\textbf{Pre-training advantage under extreme data scarcity.} On Real-World IQ (OOD), RFPrompt leads at every shot count with the largest margins at low $K$. At $K{=}2$, RFPrompt achieves $52.6\%$ versus a maximum of $18.7\%$ across all baselines, a gap of $+33.9$ percentage points. At $K{=}8$, the advantage over ResNet18, the strongest conventional model at that budget, remains $+40.5$ points. This reflects a fundamental difference in the learning problem: conventional and meta-learning baselines must discover useful representations from scratch within a few-shot budget, a task that is severely underdetermined under hardware mismatch. RFPrompt begins with pre-trained expert representations that already encode the wireless time-frequency structure; the prompt tokens need only steer the existing attention distribution toward modulation-relevant features, requiring far fewer gradient steps to achieve reliable convergence. The steady improvement from $52.6\%$ at $K{=}2$ to $83.3\%$ at $K{=}128$ confirms that prompt adaptation scales stably with supervision without the instability observed in some baselines.

% \textbf{Behavior on IEEE Dataport (in-distribution).} RFPrompt leads or ties at all shot levels on IEEE Dataport except $K{=}32$, where EfficientNet-B0 achieves $59.6\%$ versus RFPrompt's $58.8\%$. By $K{=}128$, ResNet18 ($72.8\%$) and MAML ($72.4\%$) nearly match RFPrompt ($73.0\%$), consistent with the Stage~A finding: when the evaluation distribution is well-aligned with the model's prior, the pre-training advantage shrinks as supervision grows.

RFPrompt leads or ties at all shot levels on IEEE Dataport except $K{=}32$, where EfficientNet-B0 achieves $59.6\%$ versus RFPrompt's $58.8\%$. SpectrumFM improves with the shot budget but remains below RFPrompt in this few-shot setting, reaching $46.97\%$ at $K{=}128$. By $K{=}128$, ResNet18 ($72.8\%$) and MAML ($72.4\%$) nearly match RFPrompt ($73.0\%$), consistent with the Stage~A finding: when the evaluation distribution is well-aligned with the model's prior, the pre-training advantage shrinks as supervision grows.

\vspace{-4pt}
\subsection{t-SNE Representation Analysis}

\begin{figure}[t!]
  \centering
  \includegraphics[width=\linewidth]{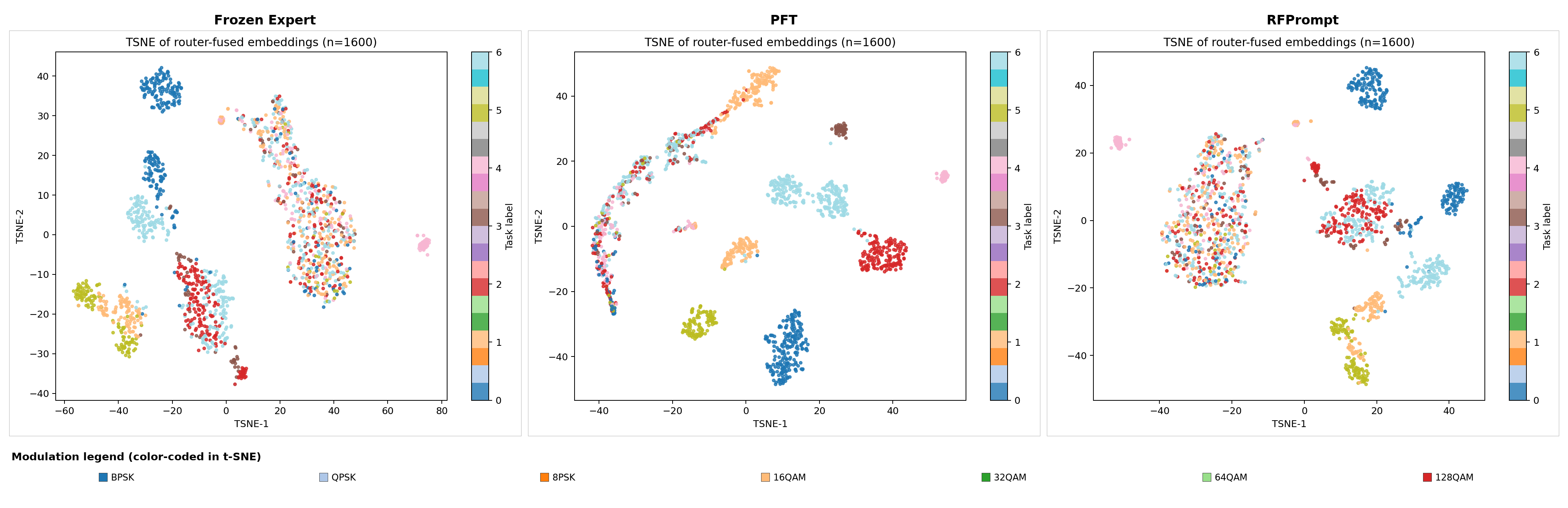}
  \caption{t-SNE projections of router-fused embeddings $\bz_{\text{final}}$ on the Real-World IQ test set ($n{=1600}$), color-coded by modulation class. 
  % \fa{can we get a better example for this plot?}
    }
  \label{fig:tsne}
\end{figure}

Fig.~\ref{fig:tsne} provides a representation-level interpretation of the Stage~A results. Frozen expert embeddings exhibit substantial class overlap in the central region, indicating that protocol-level pre-training alone does not yield strongly modulation-discriminative representations. PFT produces more spatially separated clusters, consistent with backbone weight updates that explicitly reorganize the feature space toward class boundaries. RFPrompt embeddings are visually denser and, in several regions, less globally separated than PFT embeddings, yet RFPrompt attains higher downstream accuracy across all OOD training caps with $N \geq 200$. The result is better OOD generalization under limited supervision than PFT, which reorganizes the feature space more aggressively at the cost of greater susceptibility to backbone drift.

\vspace{-4pt}
\subsection{Ablation Study: Prompt Length}

% % ── TABLE 3 ──────────────────────────────────────────────────────────
% \begin{table}[t!]
%   \centering
%   \caption{Ablation: effect of prompt length $P$ on accuracy and
%     macro F1 at extreme few-shot ($N{=}100$) and high-data
%     ($N{=}1600$) regimes.
%     \textbf{Bold} = best per column.}
%   \label{tab:ablation}
%   \setlength{\tabcolsep}{3.5pt}
%   \renewcommand{\arraystretch}{1.05}
%   \begin{tabular}{@{}clcccc@{}}
%     \toprule
%     & & \multicolumn{2}{c}{\textbf{IEEE Dataport}}
%       & \multicolumn{2}{c}{\textbf{RW-IQ}} \\
%     \cmidrule(lr){3-4}\cmidrule(lr){5-6}
%     $P$ & $N$ & Acc & F1 & Acc & F1 \\
%     \midrule
%     \multirow{2}{*}{8}
%       & 100  & 0.534 & 0.582 & 0.482 & 0.436 \\
%       & 1600 & 0.655 & 0.566 & 0.686 & 0.712 \\
%     \midrule
%     \multirow{2}{*}{12}
%       & 100  & 0.548 & 0.590 & 0.519 & 0.499 \\
%       & 1600 & 0.593 & 0.563 & 0.685 & 0.712 \\
%     \midrule
%     \multirow{2}{*}{\textbf{16}}
%       & 100  & \textbf{0.702} & \textbf{0.639}
%              & \textbf{0.689} & \textbf{0.678} \\
%       & 1600 & \textbf{0.724} & \textbf{0.696}
%              & \textbf{0.801} & 0.713 \\
%     \midrule
%     \multirow{2}{*}{20}
%       & 100  & 0.553 & 0.589 & 0.570 & 0.574 \\
%       & 1600 & 0.649 & 0.562 & 0.782 & \textbf{0.783} \\
%     \midrule
%     \multirow{2}{*}{32}
%       & 100  & 0.541 & 0.573 & 0.521 & 0.507 \\
%       & 1600 & 0.652 & 0.581 & 0.688 & 0.714 \\
%     \bottomrule
%   \end{tabular}
% \end{table}

\begin{table}[t!]
  \centering
  \caption{Ablation: prompt length $P$ vs. accuracy / macro F1. \textbf{Bold} = best per column.}
  \label{tab:ablation}
  % Reduce horizontal padding to ensure it fits one column
  \setlength{\tabcolsep}{2.8pt} 
  \renewcommand{\arraystretch}{1.1}
  \scriptsize
  \begin{tabular}{@{}l cccc cccc@{}}
    \toprule
    & \multicolumn{4}{c}{\textbf{IEEE Dataport IQ}} & \multicolumn{4}{c}{\textbf{RW-IQ}} \\
    \cmidrule(lr){2-5} \cmidrule(l){6-9}
    & \multicolumn{2}{c}{$N=100$} & \multicolumn{2}{c}{$N=1600$} & \multicolumn{2}{c}{$N=100$} & \multicolumn{2}{c}{$N=1600$} \\
    \cmidrule(lr){2-3} \cmidrule(lr){4-5} \cmidrule(lr){6-7} \cmidrule(l){8-9}
    $P$ & Acc & F1 & Acc & F1 & Acc & F1 & Acc & F1 \\
    \midrule
    8  & 0.534 & 0.582 & 0.655 & 0.566 & 0.482 & 0.436 & 0.686 & 0.712 \\
    12 & 0.548 & 0.590 & 0.593 & 0.563 & 0.519 & 0.499 & 0.685 & 0.712 \\
    % \rowcolor{gray!10} % Subtle highlight for the chosen P
    \textbf{16} & \textbf{0.702} & \textbf{0.639} & \textbf{0.724} & \textbf{0.696} & \textbf{0.689} & \textbf{0.678} & \textbf{0.801} & 0.713 \\
    20 & 0.553 & 0.589 & 0.649 & 0.562 & 0.570 & 0.574 & 0.782 & \textbf{0.783} \\
    32 & 0.541 & 0.573 & 0.652 & 0.581 & 0.521 & 0.507 & 0.688 & 0.714 \\
    \bottomrule
  \end{tabular}
  \vspace{-2mm} % Actively pulls up the following text to close gaps
\end{table}

Table~\ref{tab:ablation} sweeps prompt length $P \in \{8, 12, 16, 20, 32\}$, identifying $P{=}16$ as the global optimum for accuracy and F1 stability. While $P{=}20$ marginally improving F1 on Real-World IQ ($N{=}1600$), $P{=}16$ offers superior consistency across few-shot and in-distribution regimes. The observed non-monotonic performance suggests a capacity trade-off: insufficient tokens ($P{=}8$) lack expressive power for layer-wise modulation, whereas excessive lengths ($P{=}32$) introduce optimization noise that hinders generalization under limited label budgets.

\vspace{-4pt}
\subsection{Comparison with AMC-Specific Fine-Tuning}

% ── TABLE 4 ──────────────────────────────────────────────────────────
\begin{table}[t!]
  \centering
  \caption{Comparison with AMC-specific fine-tuning (best variant: Partial fine-tuning of the AMC backbone).}
  \label{tab:amc}
  \setlength{\tabcolsep}{3.5pt}
  \renewcommand{\arraystretch}{1.00}
  \begin{tabular}{@{}llcccccc@{}}
    \toprule
    & & \multicolumn{3}{c}{\textbf{IEEE Dataport IQ}}
      & \multicolumn{3}{c}{\textbf{RW-IQ (OOD)}} \\
    \cmidrule(lr){3-5}\cmidrule(lr){6-8}
    & \textbf{$N$}
      & \textbf{RFP} & \textbf{MAML} & \textbf{Tr}
      & \textbf{RFP} & \textbf{MAML} & \textbf{Tr} \\
    \midrule
    \multirow{5}{*}{\rotatebox[origin=c]{90}{\small Stage A}}
      & 100  & \textbf{0.70} & 0.68 & 0.67
             & \textbf{0.69} & 0.25 & 0.15 \\
      & 200  & 0.71 & \textbf{0.77} & 0.74
             & \textbf{0.79} & 0.28 & 0.16 \\
      & 400  & 0.74 & \textbf{0.80} & 0.78
             & \textbf{0.82} & 0.30 & 0.15 \\
      & 800  & 0.70 & \textbf{0.81} & 0.81
             & \textbf{0.80} & 0.33 & 0.41 \\
      & 1600 & 0.72 & 0.82 & \textbf{0.82}
             & \textbf{0.80} & 0.35 & 0.16 \\
    \midrule
    \multirow{7}{*}{\rotatebox[origin=c]{90}{\small Stage B}}
      & 2   & \textbf{37.4} & 30.6 & 33.0
            & \textbf{52.6} & 15.9 & 13.2 \\
      & 4   & \textbf{47.0} & 38.1 & 36.0
            & \textbf{60.6} & 17.7 & 13.5 \\
      & 8   & \textbf{50.2} & 41.5 & 37.4
            & \textbf{69.4} & 18.5 & 14.5 \\
      & 16  & \textbf{61.0} & 48.5 & 46.4
            & \textbf{69.8} & 22.6 & 13.5 \\
      & 32  & 58.8 & 52.6 & 51.7
            & \textbf{75.9} & 31.5 & 15.4 \\
      & 64  & \textbf{65.2} & 64.8 & 60.9
            & \textbf{80.4} & 40.5 & 15.5 \\
      & 128 & \textbf{73.0} & 70.9 & 68.2
            & \textbf{83.3} & 47.7 & 21.8 \\
    \bottomrule
    \addlinespace[2pt]
    \multicolumn{8}{@{}l}{\scriptsize
      RFP = RFPrompt,\ Tr = Transfer\_RML16.}
  \end{tabular}
\end{table}

% Table~\ref{tab:amc} compares RFPrompt against the strongest AMC-specific fine-tuning variant (partial fine-tuning) using MAML and Transfer\_RML16 as representative methods. On IEEE Dataport in Stage~A, AMC partial fine-tuning methods become competitive once $N \geq 200$, with MAML reaching $0.82$ $N{=}1600$ versus RFPrompt's $0.72$. In Stage~B on IEEE Dataport, RFPrompt leads, $K \leq 16$, but AMC configurations close the gap by $K{=}64$--$128$. On Real-World IQ, the comparison reverses decisively: no AMC partial configuration surpasses RFPrompt at any Stage~A or Stage~B budget, and Transfer\_RML16 consistently plateaus below $0.50$ across all $N$. This confirms that the OOD advantage of RFPrompt is structural: it arises from steering pre-trained LWM expert representations rather than from richer optimization procedures applied to weaker features. AMC fine-tuning can effectively exploit labeled data only when the downstream distribution is relatively benign, a condition that does not hold for the hardware and propagation conditions of the real-world IQ dataset.

Table~\ref{tab:amc} compares RFPrompt against the strongest AMC-specific fine-tuning variant (partial fine-tuning) using MAML and Transfer\_RML16 as representative methods. On IEEE Dataport in Stage~A, AMC partial fine-tuning methods become competitive once $N \geq 200$, with MAML reaching $0.82$ $N{=}1600$ versus RFPrompt's $0.72$. In Stage~B on IEEE Dataport, RFPrompt leads, $K \leq 16$, but AMC configurations close the gap by $K{=}64$--$128$. On Real-World IQ, the comparison reverses decisively: no AMC partial configuration surpasses RFPrompt at any Stage~A or Stage~B budget, and Transfer\_RML16 consistently plateaus below $0.50$ across all $N$. This confirms that the OOD advantage of RFPrompt is structural: it arises from steering pre-trained LWM expert representations rather than from richer optimization procedures applied to weaker features. AMC fine-tuning can effectively exploit labeled data only when the downstream distribution is relatively benign, a condition that does not hold for the hardware and propagation conditions of the real-world IQ dataset.

\begin{table}[!t]
  \centering
  \vspace{5pt}
  \caption{Training Efficiency Comparison: Average Wall-clock Time (s/epoch). \textbf{RFP} denotes our proposed RFPrompt method.}
  \label{tab:palmetto_timing}
  \vspace{-4pt}
  \setlength{\tabcolsep}{3.5pt} % Tighten horizontal space
  \renewcommand{\arraystretch}{1.05}
  \footnotesize % Slightly smaller font to save space
  \begin{tabular}{@{}l ccc c ccc@{}}
    \toprule
    & \multicolumn{3}{c}{\textbf{IEEE Dataport}} & & \multicolumn{3}{c}{\textbf{Real-World IQ}} \\
    \cmidrule(lr){2-4} \cmidrule(l){6-8}
    \textbf{Scale ($N$)} & \textbf{Frozen} & \textbf{PFT} & \textbf{RFP} & & \textbf{Frozen} & \textbf{PFT} & \textbf{RFP} \\
    \midrule
    100  & 33.73 & 24.52 & 24.49 & & 12.42 & 26.85 & 26.72 \\
    200  & 17.59 & 38.26 & 38.13 & & 17.50 & 45.88 & 45.64 \\
    400  & 24.39 & 63.92 & 63.90 & & 25.72 & 76.90 & 76.51 \\
    800  & 44.40 & 123.9 & 123.1 & & 48.53 & 150.6 & 149.7 \\
    1600 & 218.2 & 378.5 & 376.0 & & 256.9 & 466.2 & 465.7 \\
    \bottomrule
  \end{tabular}
  \vspace{-4pt}
\end{table}
\vspace{-4pt}
% \subsection{Training Efficiency Comparison}
% We compare computational overhead via average wall-clock training time in Table~\ref{tab:palmetto_timing}. While Frozen is faster due to zero backbone backpropagation, RFPrompt maintains nearly the same efficiency as PFT. At $N=1600$ (RW-IQ), RFPrompt requires \textbf{465.7\,s/epoch} vs. \textbf{466.2\,s/epoch} for PFT. This confirms that deep prompting enhances accuracy without meaningful latency, offering a highly efficient solution for real-time O-RAN deployment.

\vspace{-4pt}

\section{Conclusion}

This paper studied parameter-efficient adaptation of wireless foundation models for automatic modulation classification under distribution shift and limited supervision. We proposed RFPrompt, a prompt-based adaptation framework that steers a frozen pretrained backbone through learnable deep prompt tokens while updating only a small set of task-specific parameters. Rather than fully fine-tuning the backbone, RFPrompt preserves the structure learned during large-scale pretraining and adapts the model through lightweight prompt, routing, and classification components.

Using the LWM as a representative wireless foundation model, we validated RFPrompt on challenging modulation-classification benchmarks that span controlled IQ captures and real-world over-the-air recordings. Across these settings, RFPrompt demonstrates that pretrained RF representations can be adapted effectively through lightweight prompt-based steering rather than extensive backbone updates. Compared with additional baselines including SpectrumFM, RFPrompt remains strongest in the few-shot OOD regime, while SpectrumFM is most competitive in higher-data Stage~A settings. The gains are especially pronounced when labeled data are scarce, and deployment conditions differ from the pretraining environment, where RFPrompt provides robust target-domain adaptation while maintaining strong parameter efficiency. These findings suggest that prompt learning is a promising mechanism for deploying wireless foundation models in practical RF environments, where labeled data are scarce, and deployment conditions often differ from those in pretraining data.

Future work will extend this framework to investigate routing-aware and class-aware prompt design, and develop attention or attribution analyses to better explain how learned prompts modify RF time-frequency representations.

% \begin{table}[t!]
% \centering
% \caption{Average epoch wall-clock time (s/epoch) for Palmetto N-sweep runs.}
% \label{tab:palmetto_epoch_time_ieee_realworld}
% \small
% \begin{tabular}{lcccc}
% \hline
% Dataset & N  & Frozen & PFT & RFPrompt \\
% \hline
% \multirow{5}{*}{IEEE} 
%   & 100  & 33.73  & 24.52  & 24.49 \\
%   & 200  & 17.59  & 38.26  & 38.13 \\
%   & 400  & 24.39  & 63.92  & 63.90 \\
%   & 800  & 44.40  & 123.92 & 123.16 \\
%   & 1600 & 218.21 & 378.54 & 376.04 \\
% \hline
% \multirow{5}{*}{RealWorld IQ}
%   & 100  & 12.42  & 26.85  & 26.72 \\
%   & 200  & 17.50  & 45.88  & 45.64 \\
%   & 400  & 25.72  & 76.90  & 76.51 \\
%   & 800  & 48.53  & 150.60    & 149.78 \\
%   & 1600 & 256.90    & 466.25    & 465.76 \\
% \hline
% \end{tabular}
% \end{table}

\bibliographystyle{IEEEtran}
\bibliography{references}

\end{document}